# Medicines Question Answering System, MeQA

**Jesús Santamaría,**[1]

[1] Agencia Española de Medicamentos y Productos Sanitarios (AEMPS)

**Abstract:** In this paper we present the first system in Spanish capable of answering questions about medicines for human use, called MeQA (Medicines Question Answering), a project created by the Spanish Agency for Medicines and Health Products (AEMPS, for its acronym in Spanish). Online services that offer medical help have proliferated considerably, mainly due to the current pandemic situation due to COVID-19. For example, websites such as Doctoralia[1], Savia[2], or SaludOnNet[3], offer *Doctor Answers* type consultations, in which patients or users can send questions to doctors and specialists, and receive an answer in less than 24 hours. Many of the questions received are related to medicines for human use, and most can be answered through the leaflets. Therefore, a system such as MeQA capable of answering these types of questions automatically could alleviate the burden on these websites, and it would be of great use to such patients.

## 1    Introduction

The leaflets for medicinal products for human use include their complete composition and instructions for their administration, use, and storage. Adverse effects, their interactions and contraindications are also specified. In addition, the text is written clearly, and they have to pass a readability test[4] so that the number of possible lexical, syntactic, or semantic errors is very low. These characteristics make leaflets a relatively easy resource to process using Natural Language Processing (NLP) techniques.

This article presents the description of a project of the AEMPS, for the realization of a system capable of answering questions in relation to medicines for human use, called Medicines Question Answering, MeQA.

Systems capable of answering questions posed by users (Question Answering systems, QA) were born around 1960 (Phillips, 1960), and are among the first systems with some intelligence to be developed with computers.

A medicine QA system implies that it must be able to answer questions whose answers can be found in the medicine leaflets. Therefore, if the information about which the question is asked is not found in the universe of leaflets, the system should indicate that the answer has not been found, although (perhaps) it does exist. For example, when faced with a question such as *¿El ibuprofeno está contraindicado para los hipertensos?* (*Is ibuprofen contraindicated for hypertensive patients?*) the system should be able to answer, since such an answer is found in some of the leaflets of medicines whose active ingredient is Ibuprofen. However, when faced with the question, *¿el Ibuprofeno acelera*

---

[1] https://www.doctoralia.es/
[2] https://www.saludsavia.com/
[3] https://www.saludonnet.com/
[4] The readability tests must be carried out on a mandatory basis, in accordance with article 59(3) of Directive 2004/27/EC, which modifies 2001/83/EC, which establishes that "the leaflet must reflect the results of consultations with target groups of patients to guarantee their readability, clarity and ease of use", in accordance with the recommendations established in the published European Guidelines on the legibility of the leaflet and labeling of medicines for human use. Instructions on readability tests can be found at: https://www.aemps.gob.es/informa/notasInformativas/industria/2008/instrucciones-MTP.htm

*la pérdida de memoria?* (*does Ibuprofen accelerate memory loss?*) The system will not find the answer, since these leaflets do not clarify anything about this question.

Currently, there are many web pages that offer a service called "*Doctor Answers*", in which patients can send questions to doctors, and receive a response in less than 24 hours. For example, the web pages of Doctoralia[5], Savia[6], or SaludOnNet[7]. Although MeQA is a project that belongs to the AEMPS, many of these websites can benefit from its use, since, for example, they can redirect queries about medicines for human use (which are many) to MeQA, so it answer them automatically, thus alleviating the workload of these professionals.

In section 2 the state of the art in QA is reviewed, in section 3 the main features of MeQA are described, section 4 describes the evaluation process, and finally section 5 contains the conclusions and future work.

## 2     State of the Art

QA systems can be generically divided into two large groups (Jurafsky and Martin, 2000). The first group is made up of QA systems based on information retrieval, which handle a collection of texts. Given a question, the information retrieval system tries to find relevant passages, and then obtain a concrete answer by providing fragments of those passages. The second group, known as knowledge-based QA[8], builds a semantic representation of the question to a logical representation, and then these representations are used to query structured databases. An alternative approach to doing QA is to query a previously trained language model, forcing the model to answer a question solely from the information stored in its parameters. For example, in (Roberts, Raffel and Shazeer, 2020) they use a language model called T5. Although this solution is not yet complete to answer questions: for example, they don't work as well as classic models, they suffer from misinterpretation (unlike standard QA systems, for example, they currently cannot give users more context telling them which passage the answer came from). However, the study of the answer extraction from language models is an interesting area for future QA research.

Currently, QA systems have a lot of interest in the NLP community, and many other tasks have been generated, such as: long-form QA (Fan et al., 2019), where questions require a long answer; community QA, ComQA (Abujabal et al., 2019), which makes use of data sets of pairs of questions and answers created by a certain community such as Quora or Stack Overflow.

In addition, this task is also present in other fields of artificial intelligence (AI), such as image processing, called Image Question Answering, IQA (Gordon et al., 2017), or Visual Question Answering, VQA (Antol et al. ., 2015), in which the objective is to answer questions about certain elements present in photographs, such as the color of certain elements, objects, etc. It is such an active field of AI that a new task called Embodied Question Answering (Das et al., 2018) has recently been created, which consists of generating an agent at a random location in a 3D environment and asking it a question (such as *¿What color is the car?*). To answer, the agent must first intelligently navigate to explore the environment, collect the necessary visual information through first-person vision, and then answer the question (e.g. *orange*). This task combines different fields of AI such as language comprehension (LU), visual recognition, active perception, goal-based navigation, common sense reasoning, long-term memory, and conversion of language into actions. Within the biomedical domain, BioASQ (Tsatsaronis et al., 2015) has organized several QA tasks from structured data and free text, as well as the Medical Question Answering tasks organized in the TREC (Ben Abacha et al., 2017).

The proposed system, MeQA, has been developed following the paradigm of QA systems based on information retrieval and, as we will see, it combines machine learning and deep learning techniques. It has the advantage of being easily expandable to other languages, as it does not need to annotate large amounts of question-answer pairs, something that QA systems based on complete neural architectures such as encoder-decoders (also known as seq2seq) do need (Sutskever, Vinyals and Quoc, 2014), or those who perform fine-tuning on pre-trained models such as BERT (Devlin et al., 2019).

---

[5] https://www.doctoralia.es/
[6] https://www.saludsavia.com/
[7] https://www.saludonnet.com/
[8] These types of systems are also sometimes referred to as *Linked Data QA*.

## 3   MeQA Description

As previously mentioned, MeQA is a QA system based on information retrieval, the architecture of which is shown in Figure 1.

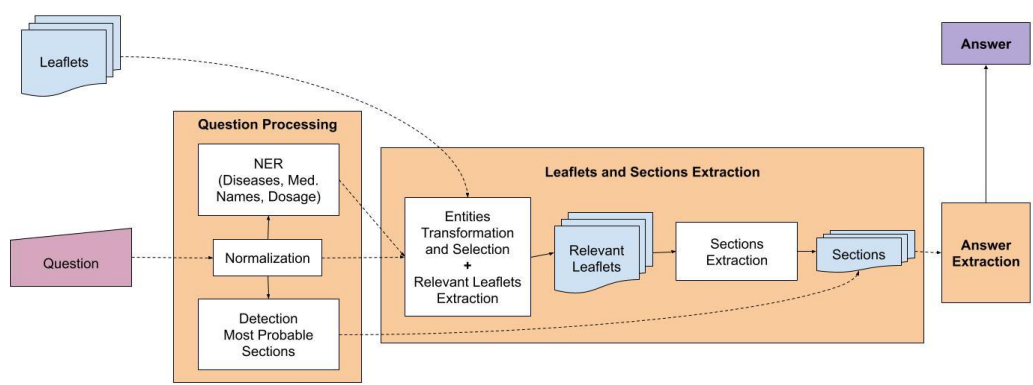

Figure 1: General architecture of MeQA.

We describe below, the three large blocks shown in figure 1.

### 3.1   Question Processing

This block is in charge of processing the user's question, extracting certain keywords, and also determines which are the most probable sections of the leaflet in which the answer will be found, along with said probability. It is made up of the following three modules:

The Normalization module takes the original question as input and performs cleaning and tokenization tasks. Specifically: convert the question to lowercase, remove irrelevant punctuation marks, remove accents, spell-correct misspelled words, and tokenize.

The NER (named entity recognizer) module takes the standardized question as input, and detects diseases, medicines names, doses, and pharmaceutical forms[9]. It is based on the maximum coincidence algorithm (Liu, 2000) with n-grams, and uses the Unified Medical Language System (UMLS, ontology containing biomedical vocabulary) as a dictionary of diseases (Bodenreider, 2004). For the names of medicines, the doses, and the pharmaceutical forms, the CIMA REST[10] interface is used, which allows the query of the information of the medicines authorized by AEMPS. The extraction of the pharmaceutical form of the medicine is extracted whether it is named directly or not in the question. To do this, the dependency tree[11] of the normalized question is generated, and the *head* element of the medicine name is obtained, thus obtaining the possible pharmaceutical forms for the named medicine. Figure 2 shows a fragment of the dependency tree generated with SpaCy[12] for the normalized question: *una persona mayor con la tensión alta puede tomar ibuprofeno 600 mg gracias* (*an elderly person with high blood pressure can take ibuprofen 600 mg thanks*).

---

[9] The pharmaceutical form (also called galenic form) is the individualized arrangement to which medicinal substances and excipients are adapted to constitute a medicine. The most common pharmaceutical forms are those that are administered orally, such as tablets, oral solutions, etc.

[10] https://www.aemps.gob.es/apps/cima/docs/CIMA_REST_API.pdf

[11] The internal structure of a dependency parser consists solely of directed relationships between lexical items in the sentence. The arguments for these relationships consist of an element named *head* and another named *dependent*.

[12] https://spacy.io/

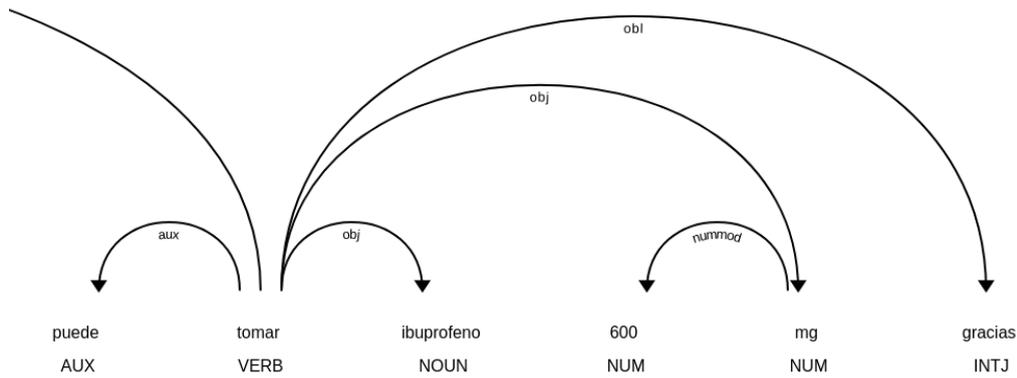

Figure 2: Fragment of the dependency tree for the sentence: *una persona mayor con la tensión alta puede tomar ibuprofeno 600 mg gracias* (*an elderly person with high blood pressure can take ibuprofen 600 mg thanks*).

In this figure it can be seen that ibuprofeno (ibuprofen) is *dependent* on tomar (take) and, therefore, the pharmaceutical forms of ibuprofeno (ibuprofen) must be related to tomar (take), such as tablets, or soft capsules, while pharmaceutical forms such as gel or injectable solution will be discarded, because they are not related to the verb tomar (take).

Finally, the model that determines the most probable sections in which the answer will be found is shown in Figure 3.

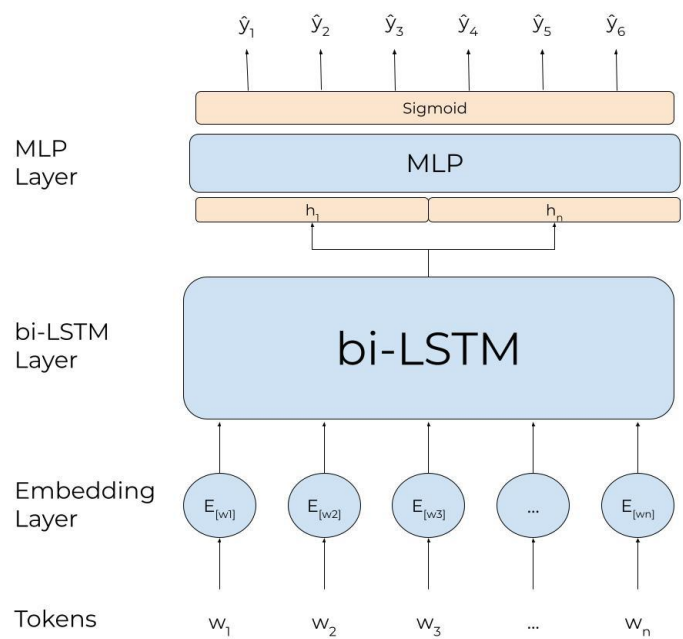

Figure 3: Architecture of the model that determines the most probable sections in which the answer will be found.

The tokens of the question are converted into embeddings (learned during the training of the model itself), and they are passed to the bidirectional LSTM layer (Bidirectional Long Short-Term Memory, bi-LSTM; Hochreiter and Schmidhuber, 1997), this layer connects directly with a fully connected layer (multilayer perceptron, MLP) with sigmoid activation function and 6 outputs (one per section). To summarize the architecture:

$$\hat{y}_i = \sigma(\text{MLP}([\text{LSTM}^f(x_{1:n}); \text{LSTM}^b(x_{n:1})]))$$

$$x_{1:n} = E_{[w_1]}, \cdots, E_{[w_n]}$$

The hyperparameters of this model are the following:
- Number of neurons in the bi-LSTM layer: 32 for the forward and backward layers, total 64
- Number of neurons in the MLP layer: 6
- Dropout: 0.5
- Recurrent Dropout: 0.5
- Loss function: binary cross entropy
- Optimizer: ADAM (Kingma, D. and Ba, J., 2014), learning rate: 0.001
- Number of Epochs: 4

To train this network, a corpus of 13,989 questions on medicines for human use has been generated, extracted from the Doctoralia website, annotated with the section or sections in which the answer is found[13]. Other text classification approaches have been tested for this module, such as naive Bayes or logistic regression, and the performance obtained is quite close to that obtained by the proposed architecture when the size of the annotated data set is smaller[14]. As the size is increased, the proposed architecture performs better.

## 3.2 Leaflets and Sections Extraction

This block extracts the most relevant leaflets and relevant sections of said leaflets. It is made up of two modules:

The module for selection of leaflets and entities transformation, takes as input the entities extracted by NER (diseases, medicines names, doses, and pharmaceutical forms), together with the set of all leaflets, and uses a unsupervised vector representation model to determine which is the most relevant leaflet, with respect to the diseases extracted and expanded through similar concepts with UMLS[15] through the TF-IDF metric, from among all those leaflets that share the name of the extracted medicine, the dose, and the pharmaceutical form.

The section extraction module predicts additional sections (to the main sections predicted by bi-LSTM), which may contain relevant information. It is composed of a vector space model (VSM), based on TF-IDF, and a LSI (Latent Semantic Indexing) topic model (Dumais, 2005) with 6 topics. The reason for using VSM and LSI is because the architecture based on bi-LSTM only predicts in which sections the answer will be found, and therefore, any additional information that might be relevant to the patient, but that is outside the predicted sections, would not be captured.

## 3.3 Answer Extraction

This block *builds* the answer from the information provided by the previous blocks. The tasks it performs are the following:

Extract sentences from the selected leaflets for each of the predicted sections, remove duplicate sentences, add context to each of the sentences (sections or list beginning, where the sentences are located) so that the extracted information is more understandable, group the information by sections, and classify by relevance. In this way, as already mentioned, the information extracted by the section predicted by the bi-LSTM network will be the response, while the information extracted by the sections predicted by VSM and by LSI will be shown as additional information.

---

[13] All leaflets have the same structure (content, what it is and what it is used for, what you need to know, how to use/take, adverse effects, storage, and composition), so it takes a trained person about 2 weeks to make the annotation of the 13,989 questions.

[14] In fact, both naive Bayes and logistic regression perform better than the proposed architecture when the size of the data set is less than 5,000 annotated questions.

[15] In UMLS there is a field called CUI (Concept Unique Identifier) that links strings with the same meaning. Its definition can be consulted in the CUI entry, in the glossary: https://www.nlm.nih.gov/research/umls/new_users/glossary.html

Figure 3 shows the answer and additional information for the question: *tengo reflujo por perforación gastrointestinal, ¿puedo tomar cidine?* (*I have reflux due to a gastrointestinal perforation, can I take cidine?*).

Figure 3: Answer given by MeQA to the question: *tengo reflujo por perforación gastrointestinal, ¿puedo tomar cidine?* (*I have reflux due to a gastrointestinal perforation, can I take cidine?*). In the figure you can distinguish the answer and the additional information. In addition, the detected UMLS concepts and context elements are displayed.

As can be seen in Figure 3, the answer is in section 2 (predicted by bi-LSTM), while the additional information is in section 1 (predicted in this case by LSI). This figure has highlighted the additional context elements that have been added (*No tome Cidine 1 mg/5 ml Solución oral - Do not take Cidine 1 mg/5 ml Oral solution*), as well as the UMLS concepts detected in the leaflet and related to the question (*perforación gastrointestinal* and *reflujo gastroesofágico - gastrointestinal perforation* and *gastroesophageal reflux*).

## 4     MeQA Evaluation

MeQA has been evaluated automatically and manually. Automatic evaluation allows to know the performance of MeQA, while manual evaluation seeks to detect certain biases that MeQA could have, and that cannot be detected by automatic evaluation. Chapter 2 of (Clark, Fox, and Lappin, 2010) highlights the importance of conducting both types of evaluations.

### 4.1     Automatic Evaluation

For the automatic evaluation, a corpus of 300 questions on medicines for human use has been created, randomly selected from the Doctoralia website. In order to evaluate the general performance of MeQA, as well as each of its modules described above, the fields that have been annotated, for each question, in the leaflets have been: medicines & sections, answer, and reference number[16]. This annotation process, as it is to be expected, is much more complex than the one described above to annotate only the sections in which the answer is found, and therefore, it has also been necessary to invest more time to carry it out[17]. An example is shown in Figure 4, with the entities annotated: *question, medications & sections, answer, reference number*.

---

[16] Reference numbers allow you to uniquely identify medicine leaflets.
[17] It took about 3 months to annotate these 300 questions.

| Question | Medications ||| Sections | Answer | Reference Number |
|---|---|---|---|
| Tengo el pie y tobillo hinchado y enrojecido por que extrajeron el yeso a causa de una fractura de tibia y peroné. Puedo usar pomada hirudoid forte para tratarlo? | HIRUDOID|||1 | HIRUDOID|||1: Este medicamento está indicado para: el alivio local sintomático de los trastornos venosos superficiales como pesadez y tirantez en piernas con varices en adultos. El alivio local sintomático de hematomas superficiales producidos por golpes en adultos y niños mayores de 1 año. | 58289 |
| ... | ... | ... | ... |

Figure 4: Example of annotated entities for the evaluation corpus question: *Tengo el pie y tobillo hinchado y enrojecido por que extrajeron el yeso a causa de una fractura de tibia y peroné. Puedo usar pomada hirudoid forte para tratarlo?* (*My foot and ankle are swollen and red because the cast was removed due to a fracture of the tibia and fibula. Can I use hirudoid forte ointment to treat it?*)

The annotation process that has been followed is adapted from (Roberts, 2009) and is shown in Figure 5.

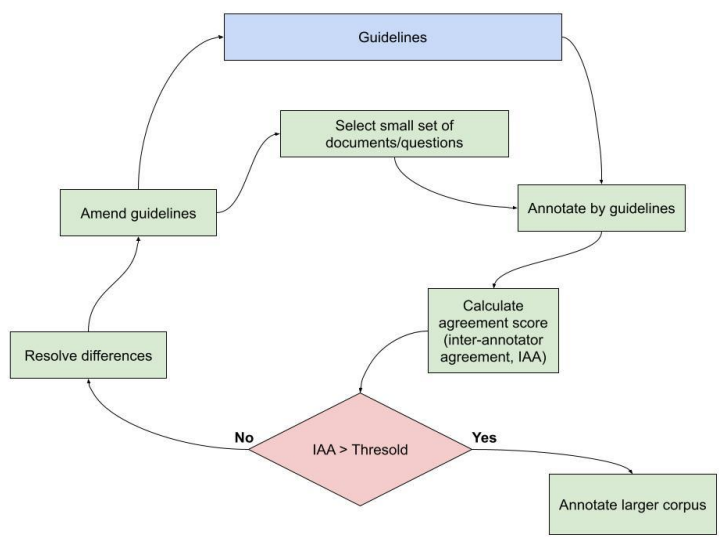

Figure 5: Annotation process followed to create the evaluation corpus.

It has been developed by experts in medical data from the AEMPS. The process shown in the figure is as follows: you start with a draft of the annotation guide, and with a random set of 50 questions. Once annotated, following the guidelines of the annotation guide, the differences between the annotations are compared, and these differences are calculated using the IAA score (inter-annotator agreement). If the IAA is greater than or equal to 95%, the rest of the corpus is annotated, but if not, the differences are reflected in the annotation guide, and another 50 questions are obtained to annotate, which starts a new one iteration. In our case, 5 iterations have been carried out to reach 95% agreement.

Since MeQA does not return a list of responses, but only one, the Mean Reciprocal Rank (MRR) metric introduced in the TREC Q/A track from 1999 (Voorhees, 1999) (Fukumoto, Kato, and Masui, 2004) has not been used. Instead, (Rajpurkar et al., 2016) has been followed. For reading comprehension systems (Hirschman et al., 1999) in datasets like SQuAD, two metrics are often used, both of which ignore punctuation marks and articles (*el, la, los, las, un, una, unos, unas - the, a, an*):
- Exact Match: The percentage of predicted answers that match the gold answer exactly.
- $F_1$: The average word/token overlap between predicted and gold answers. Treat the prediction and gold as a bag of tokens, and compute $F_1$ for each question, then return the average $F_1$ over all questions.

The results obtained by MeQA, by modules, measured by $F_1$, and by end-to-end[18] evaluation, have been the following:
- NER: 97%
- bi-LSTM: 91%
- Leaflets extraction: 97%
- Sections extraction: 82%
- Answer extraction: 87%

The final performance of MeQA, therefore, corresponds to the module of answer extraction (the information that is delivered to the user), and that is 87%.

The questions in the evaluation corpus have been analyzed, classified into three categories, and the following conclusions have been drawn:
1. Presence of very complex sentences: These are phrases that are difficult to understand, even for a human being. The complexity is given by its length (they are very long sentences), or by its interpretation (it can have different meanings). An example sentence of the first type is the following:

    *Buenas tardes, tengo 50 años, estoy tomando Duodart desde hace 8 meses por padecer HBP, y desde entonces, a parte de disminuir mi deseo sexual ( que por lo visto es normal) me encuentro muy cansado, ( la analítica de sangre da valores normales en todo) y además al día siguiente de hacer cualquier ejercicio físico, parece que me han dado una paliza, me duele todo el cuerpo. esto es normal con este medicamento? estoy optando por operarme de la HBP, ya que no puedo seguir con este cansancio[19].*

    On the other hand, an example of a complex question for its interpretation is the following:

    *Cuanto tiempo puedo mantener un bote de pectox abierto?[20]*

    The question does not clarify whether it refers to expiration, loss of effectiveness, its properties, or another issue. The answer given by the Doctoralia doctor is the following:
    *El Pectox solucion es un mucolitico y se puede tener abierto no mas de tres meses sin que pierda sus propiedades (The Pectox solution is a mucolytic and can be kept open for no more than three months without losing its properties).*
    However, such information is not in the content of the Pectox leaflet, hence the answer is *No Answer*. As you might expect, it is with these types of phrases that MeQA performs the worst.

---

[18] Most NLP systems today are not monolithic entities, but consist of different components, often arranged in a pipeline. For example, a parser depends on the POS tagger, which in turn depends on tokenization, which in turn depends on the sentence splitter. In the end-to-end evaluation of this parser that requires POS tags as input, the quality of the parse trees obtained is evaluated based on the output of a real POS tagger which, logically, may contain errors. End-to-end evaluations provide a meaningful quantification of the effectiveness of the system in real-world circumstances.

[19] *Good afternoon, I am 50 years old, I have been taking Duodart for 8 months due to BPH, and since then, apart from reducing my sexual desire (which apparently is normal) I am very tired, (the blood test gives normal values in everything) and also the day after doing any physical exercise, it seems that I have been beaten, my whole body hurts. is this normal with this medicine? I am opting to have BPH surgery, as I cannot continue with this fatigue.*

[20] *How long can I keep a bottle of pectox open?*

2. Presence of complex sentences: They are sentences of medium length, ungrammatical sentences, and even with a certain subjectivity. An example of a question of medium length and ungrammatical is the following:

   *Mi hijo toma tryptizol de 25 mg y escitalopram de 10 mg ahora por somnolencia matutina le bajaron a 10 mg pero parece que le esta dando un bajón y que esta mas nervioso, ¿Será de bajar el tryptizol, seria el que realmente le esta haciendo bien en el tratamiento, y el escitalopram sueño?[21]*

   On the other hand, a phrase that has a certain subjectivity is the following:

   *¿Presentar taquicardia en tratamiento con Ventolin, es motivo para suspender la administración?[22]*

   The correct answer, given through the information that comes in the leaflet, is an indirect answer, since it is not indicated (in the leaflet) if the administration should be suspended when presenting tachycardia. The answer given by MeQA is an accurate but incomplete answer. MeQA get good results with these types of phrases.
3. Presence of normal sentences: These are short and direct phrases which, a priori, can be answered with the information from the leaflet. Examples of these types of questions are the following:

   *¿Medebiotin Fuerte se puede tomar con hipertensión?*
   *Por favor, ¿Puede causarme hipoglucemias la Olanzapina?*
   *¿Hemicraneal sube la tensión ocular?[23]*

   In this type of sentences MeQA obtains the best results.

Most of the questions present in the evaluation corpus are complex questions, corresponding to the second type, hence the results obtained by MeQA are good, but can be improved.

## 4.2 Manual Evaluation

The manual evaluation was carried out with 13 members of the Division of Pharmacology and Clinical Evaluation of the AEMPS. Figure 6 shows the interface that has been provided to the evaluators.

---

[21] *My son takes tryptizole 25 mg and escitalopram 10 mg now because of morning drowsiness they lowered him to 10 mg but it seems that he is giving a slump and that he is more nervous, will it be to lower the tryptizole, would it be the one that is really doing him well in treatment, and sleep escitalopram?*
[22] *Is the presence of tachycardia in treatment with Ventolin a reason to suspend the administration?*
[23] *Can Medebiotin Forte be taken with hypertension?*
*Please, can Olanzapine cause hypoglycemia?*
*Does hemicranial increase ocular tension?*

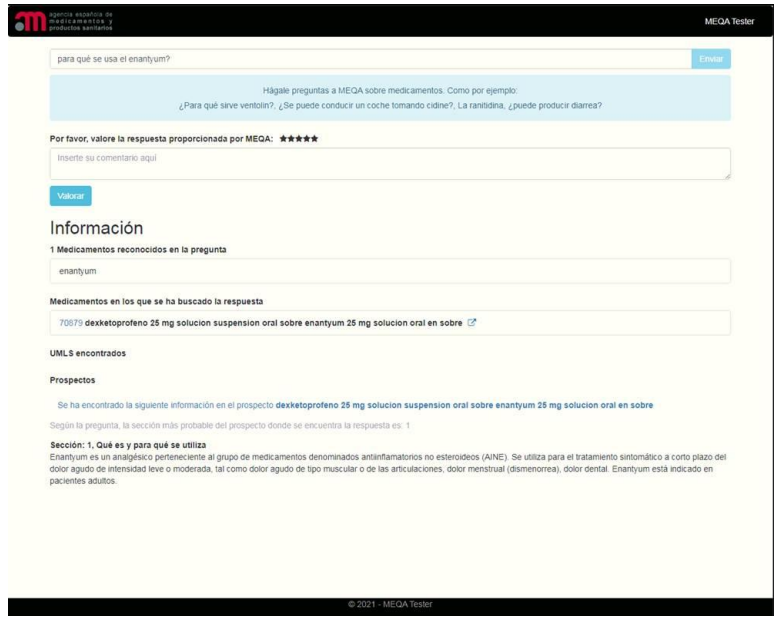

Figure 6: Interface provided to the evaluators AEMPS team.

The upper part of the interface allows you to evaluate each MeQA response by a 5-star rating, and also to write comments on that response. The number of questions that have been asked, and therefore evaluated, manually have been 334, of which 205 do not contain comments. The comments made on the 129 questions have been classified into 5 different categories shown in figure 7, along with the number of comments made by category.

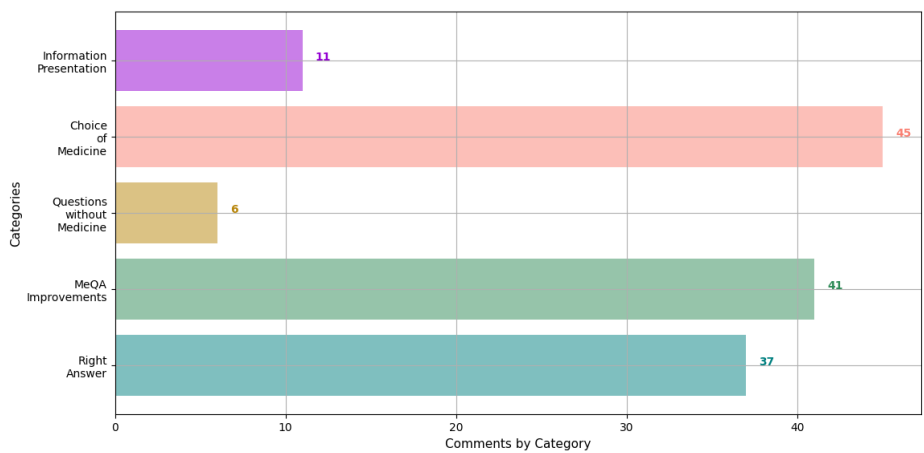

Figure 7: Classification in 5 categories of the comments made by the evaluators, together with the number of comments made by category.

Figure 8 shows some examples of comments for each of the categories, and whether or not they have generated action points.

| Comment Category | Examples | Action Points |
|---|---|---|
| Information Presentation | Information should be more separated<br>Organize it to make it more read friendly<br>Highlight keywords more | ✗ |
| Choice of Medicine | Use only marketed medicines<br>Refers to aspirin complex, which is not a mono-component<br>Dose selected is not the most usual | ✓ |
| Questions without Medicine | The use of MeQA can be undermined<br>It can become a virtual Doctor<br>May enhance self-medication | ✓ |
| MeQA Improvements | First sentence should be deleted<br>Duplicate information<br>Some sentences have not been selected | ✓ |
| Right Answer | Perfectly answers my question<br>Perfect answer<br>Okay | ✗ |

Figure 8: Examples of comments for each of the categories, and whether or not they have generated action points.

The *information presentation* refers to how the information should be presented to the user so that it is easier to read. Because the information provided by MeQA to the reviewers is only a prototype, these comments have not generated any action point, and will be considered later when the final version is made.

The last category, *right answer*, logically, has not generated any element of action.

The other three categories (*choice of medicine, question without medicine, MeQA improvements*) have indeed generated action items, and they deal with the medicine that MeQA should choose; whether or not to answer questions in which no medicine is named, such as "what can I take for fatigue?"; and those related to improving MeQA such as remove duplicate information.

Some of the action points have already been implemented for these three categories, listed below, while others are in progress:

- Choice of Medicine:
  ° Don't answer questions about non-marketed medicines: Implemented.
  ° Answer with a mono-component[24] if the question reflects it: in progress.
  ° Possibility to select the dose: in progress.
- Question without Medicine:
  ° Don't answer questions that don't mention any medicine: implemented.
- MeQA Improvements:
  ° Remove duplicate sentences: implemented.
  ° Improve recall, getting missing sentences: in progress.
  ° Improve precision, removing excess sentences: in progress.

---

[24] A mono-component is a medicine that contains only one active substance.

## 5    Conclusions and Future Work

This paper describes a system developed at the AEMPS, called MeQA, which allows answering questions about medicines through the leaflet. Its architecture has been shown and explained in a general way, and also the modules that compose it.

MeQA can be very useful for users, since there are a large number of web pages offering a service called "Doctor Answers", in which most of the questions deal with medicines for human use that can be answered through the leaflet.

MeQA has been evaluated both automatically and manually. The automatic evaluation has been carried out in a general way as well as of each of the described modules, obtaining, in general, an F1 performance of 87%. MeQA combines machine learning and deep learning methods, and although it is not a semi-supervised system, it is a low-supervision system. We believe that this is precisely the great advantage of MeQA over other approaches based entirely on deep learning. MeQA hardly needs annotated data to work, only the module that predicts the sections in which the answer is likely to be found uses annotated data, the rest of the modules are unsupervised. As mentioned, the annotation of this information is very simple, and very fast, but not the complete annotation of the answer, which would be needed to develop systems based entirely on deep learning. In the future, improvements are expected to increase the performance of the system. In particular, because the leaflets are divided into sections, MeQA is able to go directly to the predicted sections. However, it would be useful to have a module that allows to go through the entire leaflet (without splitting) and obtain those sections. As explained above, MeQA performs worst on very complex questions. A possible solution to assess would consist of dividing these questions into fragments, analyzing and answering each one of them, and combining the answers into one.

## *Acknowledgments*

This project has been carried out and financed by the Spanish Agency for Medicines and Health Products, AEMPS, in the Information Systems Division. Finally, we would particularly like to thank the work team that has participated in the development of MeQA: José Manuel Simarro, Gianluca Risi, Sergio Baños, Ana López de la Rica, and Kine Toure.